\documentclass{article}

\usepackage{microtype}
\usepackage{graphicx}
\usepackage{subfigure}
\usepackage{booktabs} 
\usepackage[most]{tcolorbox}

\usepackage{hyperref}

\usepackage[preprint]{icml2025-preprint}

\usepackage{amsmath}
\usepackage{amssymb}
\usepackage{mathtools}
\usepackage{amsthm}

\usepackage[capitalize,noabbrev]{cleveref}

\theoremstyle{plain}

\theoremstyle{definition}

\theoremstyle{remark}

\usepackage[textsize=tiny]{todonotes}

\newcommand\papertitle{You Are What You Eat - AI Alignment Requires Understanding How Data Shapes Structure and Generalisation}

\icmltitlerunning{\papertitle}

\begin{document}

\twocolumn[
\icmltitle{\papertitle}



\icmlsetsymbol{equal}{*}

\begin{icmlauthorlist}
\icmlauthor{Simon Pepin Lehalleur}{equal,amsterdam}
\icmlauthor{Jesse Hoogland}{equal,timaeus}
\icmlauthor{Matthew Farrugia-Roberts}{equal,oxford}
\icmlauthor{Susan Wei}{monash}
\icmlauthor{Alexander Gietelink Oldenziel}{timaeus,ucl}
\icmlauthor{George Wang}{timaeus}
\icmlauthor{Liam Carroll}{timaeus,gradient}
\icmlauthor{Daniel Murfet}{uom}

\end{icmlauthorlist}

\icmlaffiliation{ucl}{Department of Computer Science, University College London}
\icmlaffiliation{amsterdam}{Korteweg-de Vries Institute for Mathematics, University of Amsterdam}
\icmlaffiliation{timaeus}{Timaeus}
\icmlaffiliation{oxford}{Department of Computer Science, University of Oxford}
\icmlaffiliation{monash}{Department of Econometrics and Business Statistics, Monash University}
\icmlaffiliation{uom}{School of Mathematics and Statistics, the University of Melbourne}
\icmlaffiliation{gradient}{Gradient Institute}

\icmlcorrespondingauthor{Daniel Murfet}{daniel.murfet@gmail.com}

\icmlkeywords{Machine Learning, ICML}

\vskip 0.3in
]



\printAffiliationsAndNotice{\icmlEqualContribution} 

\begin{abstract}
In this position paper, we argue that understanding the relation between structure in the data distribution and structure in trained models is central to AI alignment. First, we discuss how two neural networks can have equivalent performance on the training set but compute their outputs in essentially different ways and thus generalise differently. For this reason, standard testing and evaluation are insufficient for obtaining assurances of safety for widely deployed generally intelligent systems. We argue that to progress beyond evaluation to a robust mathematical science of AI alignment, we need to develop statistical foundations for an understanding of the relation between structure in the data distribution, internal structure in models, and how these structures underlie generalisation.
\end{abstract}

\section{Introduction}

The alignment problem was stated by the mathematician \citet{wiener1960some} as follows: ``If we use, to achieve our purposes, a mechanical agency with whose operation we cannot interfere effectively... we had better be quite sure that the purpose put into the machine is the purpose which we really desire.''
What is it that we are putting into our machines, and how sure are we that it is instilling in them our desired purposes? 
In the era of large-scale deep learning, AI alignment techniques rely on shaping the training data distribution. In turn, this data distribution shapes the internal structures that develop inside the model. These internal structures determine the model's behaviour, including its alignment properties.
In this paper, we argue for the position that \textbf{achieving a deep scientific understanding of this pipeline -- from data, to internal structure, through to generalisation -- is necessary for AI alignment.}

The paper is organised as follows:
\begin{itemize}
    \item
        In \cref{section:s4}, we outline what is known about how internal structure in neural networks emerges over training, how this structure is shaped by patterns in the data distribution, and how this structure plays a role in determining the model's generalisation behaviour.
    \item 
        In \cref{section:alignment}, we describe the alignment problem\footnotemark{} and we argue that most modern alignment techniques function by shaping the training data distribution (thus, \emph{indirectly} shaping structure and generalisation).
    \item 
        In \cref{section:slt}, we review what we understand about how data shapes structure in learning systems. We recall the theory of internal model selection in Bayesian inference with singular model classes, which predicts that in some cases, a simpler, less accurate model can be preferred over an accurate but more complex model.
    \item
        In \cref{section:risks}, we argue that in the context of attempting to align powerful AI models to a complex system of human values, this kind of inductive bias can cause a problem since many data distributions may lead to simplistic and dangerous alternative solutions.
    \item 
        In \cref{section:understanding}, we defend our position: If you want to solve the alignment problem by shaping the data distribution, you need a deep understanding of how data shapes structure and how structure determines generalisation. We sketch several research avenues that would contribute towards this goal.
\end{itemize}
We conclude by considering two alternative views (\cref{section:alt}): a structural understanding of generalisation may be unnecessary for alignment, and may be intractable. We nevertheless argue that, as during the advent of previous powerful and dangerous technologies, we should not expect to get away safely with only a shallow, empirical understanding of how AI systems develop. Moreover, while the project we outline is sure to be a challenging enterprise, this does not mean it is not necessary.

\footnotetext{%
This paper does not debate the importance of AI alignment or safety; it presumes the reader recognises these as critical issues. Instead, we focus on the need for new theoretical paradigms to address core challenges in AI alignment.
}

\begin{figure*}[ht!]
    \centering
    \includegraphics[width=\linewidth]{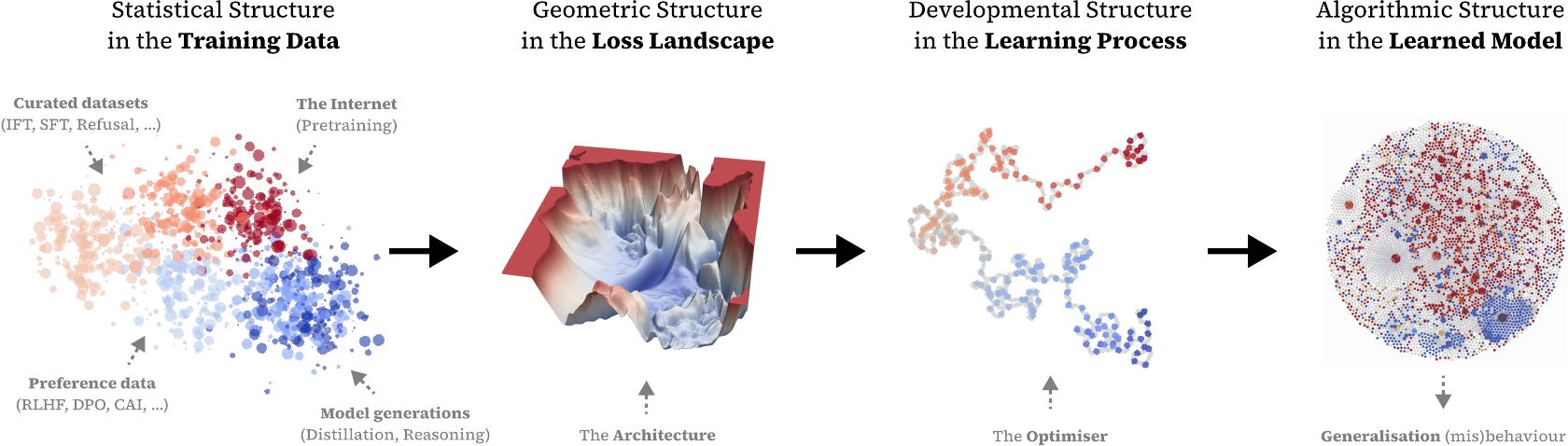}
    \caption{\textbf{From data to model behaviour}: Structure in data determines internal structure in models and thus generalisation. Current approaches to alignment work by shaping the training distribution (left), which only indirectly determines model structure (right) through the effects on shaping the optimisation process (middle left \& right). To mitigate the limitations of this indirect approach, alignment requires a better understanding of these intermediate links (loss visualisation from \citealt{li2018visualizing}; ``S4 correspondence'' based on~\citealt{wang2024differentiation}).}
    \label{fig:position}
\end{figure*}

\section{Patterns, Structures, and Generalisation}\label{section:s4}

In this section, we outline how neural networks display emergent internal structure, how this structure is shaped by patterns in the data distribution, and the relation to generalisation.
This process is summarised in \cref{fig:position}, which describes the flow of structure from the training data through the geometry of the loss landscape and the unfolding of the training process into the internal structure of the model (building on the ``S4 correspondence'' of \citealt{wang2024differentiation}).

\subsection{Internal Structure in Models}

\begin{quote}
``Understanding data and understanding models that work on that data are intimately linked. In fact, I think that understanding your model has to imply understanding the data it works on.''
\begin{flushright}
--- Chris Olah, Visualizing Representations,  \citeyear{olah2015visualizing}
\end{flushright}
\end{quote}

In classical statistics, parametrised statistical models are small, and the structure they possess is the structure that we give them (e.g., in a graphical model). In the modern era of machine learning, models are large and initially unorganised and progressively acquire ``structure'' throughout training~\citep{nanda2023progress,wang2024differentiation,tigges2024llm}.

Research in mechanistic interpretability has begun to uncover interesting internal algorithmic structure in neural networks, such as ``circuits'' for various tasks \citep{olah2020zoom,cammarata2020thread,sharkey2025open}. Researchers have identified circuits for indirect object identification \citep{wang2022interpretability}, activation patterns corresponding to individuals, emotions, and programming concepts \citep{templeton2024scaling}, and even structures that seem to implement mathematical operations \citep{nanda2023progress}. These findings suggest that models develop internal representations and algorithms that go beyond simple pattern matching or memorisation. That is not to say that we expect algorithms in neural networks to be as ``clean'' as traditional programs; we expect these learned programs to often behave like ensembles of partial programs and heuristics. While we've made progress in identifying these learned algorithms, the connection between this internal structure and generalisation remains to be mathematically clarified \citep{olsson2022context,he2024learning}. 

From a fundamental point of view, it is not surprising that neural networks acquire algorithmic structure: modern deep learning systems achieve low loss on their training objectives by effectively compressing the data distribution \citep{shwartz2017opening,deletang2023language}, and a useful form of compression is learning to represent the underlying generative process \citep{schmidhuber1992learning,bengio2013representation,ha2018world}. Learning as compression is also the key idea of the minimum description length principle (MDL), which is equivalent in some regimes to Bayesian inference~\citep{grunwald2019minimum}.  

\subsection{Structure in the World}

The concept of structure in the world as a precondition for learning has deep roots in philosophy and science~\citep{timaeus,bacon}. 
Classical statistics (e.g., \citealt{cox2006principles}) formalises this notion through the idea of a ``true" model -- a specific probability distribution believed to have generated the observed dataset. In this paradigm, learning involves using the data to accurately estimate the parameters of this true probability distribution. 
With the true model, prediction is a straightforward, subsidiary task.

However, the reliance on a true model in inferential statistics has faced long-standing criticism \citep{rissanen1998stochastic,breiman2001statistical,cox2006principles}. The advent of modern deep learning has intensified this scrutiny~\citep{sejnowski2020unreasonable}. In deep learning, a model's predictive performance is valued over its veracity.

Given deep learning's empirical success, we are compelled to re-examine what it means to learn parameters from data -- in this case, neural network weights from training data. Notably, no one believes that a CNN trained on CIFAR is learning the precise underlying probability distribution of the images. Furthermore, due to a lack of identifiability, 
multiple neural network weight configurations can yield the same loss -- and thus the same likelihood -- although they may not perform equivalently in other respects. This raises a critical question: If there is no single true parameter and many parameters result in equivalent loss, what exactly are we learning when we optimise neural network weights?

We posit that the success of deep learning in modelling complex phenomena suggests a form of structure in the world that is \textbf{algorithmic} in nature.
In this perspective, algorithms themselves can be viewed as latent features of the world. This view aligns with classic work on inductive inference in algorithmic information theory \citep{solomonoff1964formal} as well as more recent work in computational mechanics \citep{crutchfield2012between}, which propose that the most concise description of a process is often algorithmic and that the simpler algorithms realising a given process are more likely.

This is consistent with decades of research in cognitive neuroscience. For example, \citet{rogers2004semantic} argue that a natural explanation for internal structure in biological and artificial neural networks is that it mirrors internal structure in the data. Further, they argue for a relationship between these two kinds of structure and structure in the learning process, such as developmental stages. Recent work in deep learning \citep{saxe2019mathematical,hoogland2024developmental,wang2024differentiation} has provided evidence supporting the applicability of this view to artificial neural networks.

The emergence of coherent computational structures across different scales \citep{lieberum2023doescircuitanalysisinterpretability} further support the idea that these learned algorithms may reflect intrinsic properties of the world rather than artefacts of specific models \citep{huh2024platonic}, although see \citet{chughtai2023toymodeluniversalityreverse}. In the alignment discourse, a related point of view has been articulated by Wentworth; see \citet{natabs2023}. This viewpoint not only offers a lens for understanding deep learning but also calls for a re-evaluation of classical statistics in light of these algorithmic forms of structure implicit in data.

\subsection{Underspecification and Generalisation}\label{sec:underspecification}

\begin{quote}
``However, in singular learning machines, the role of the parameter is not uniquely determined, hence even by checking the parameter, we cannot examine whether the model’s learning is subject to the designer’s intention.''
\begin{flushright}
--- Sumio Watanabe, \citeyear[\S 3.4]{watanabe2024reviewprospectalgebraicresearch}

\end{flushright}
\end{quote}

The modern era of deep learning is characterised by a focus on scaling up model size, dataset size, and training compute as the principal engine of capability improvement. The belief in this approach, often referred to as the ``scaling hypothesis,'' was articulated and popularised by \citet{gwern2020scaling} based on earlier observations \citep{sutton2019bitter, amodei2018ai, raffel2020exploring}. The scaling hypothesis suggests that many AI capabilities might emerge primarily as a consequence of increasing scale rather than through fundamental algorithmic breakthroughs. This idea gained empirical support through work on scaling laws \citep{hestness2017deep, kaplan2020scaling} and the success of GPT-3 \citep{brown2020language,bommasani2021opportunities}.

In this paradigm, AI development resembles industrial chemical synthesis. Just as chemical engineers carefully control reactants, catalysts, and conditions to yield desired products, as predicted by reaction kinetics, AI engineers manipulate training data distributions, model architectures, and optimisation processes to produce models with targeted test losses, as predicted by scaling laws. However, unlike industrial chemical synthesis, the exact properties and capabilities of the resulting systems are not known in advance.

One reason for this uncertainty, beyond our poor understanding of SGD dynamics, is \textit{underspecification}~\citep{d2022underspecification}, where the training distribution does not specify a unique solution.
There can be a large set of low-loss parameters that all satisfy the constraints enforced by the loss function equally well \citep{pmlr-v162-yang22k,pmlr-v235-reizinger24a,papyan2019measurements,hochreiter1997flat,fort2019emergent,chen2024stochasticcollapsegradientnoise}. Among these low-loss parameters, there may be solutions that \emph{solve the problem in qualitatively different ways} or, in other words, that \emph{generalise differently} outside of the training distribution. 

It is our position that these differences in generalisation arise from differences in internal algorithms and representations. Thus, if we care about what happens when the model is deployed in new contexts, then we must study the structure that the model has learned and not merely its performance on the training set or evaluations \citep[\S 3.4]{watanabe2024reviewprospectalgebraicresearch}. We ideally want AI training to qualify as an \emph{engineering} process, where we have sufficient control over the behaviour of the final product in its intended environment, even if this differs from the training environment.

\section{AI Alignment by Data Distribution Shaping}\label{section:alignment}

In this section, we describe what we mean by alignment and then review state-of-the-art methods for AI alignment, most of which function by shaping the data distribution so as to \emph{indirectly program model behaviour.}

\subsection{The Alignment Problem}

While concerns about alignment have existed since the beginning of computer science \citep{turing1948, good1965}, the discourse gained momentum in the early 21st century \citep{yudkowsky2004coherent,bostrom2014superintelligence}. Recent progress in artificial intelligence has brought a new focus to AI misalignment as a source of profound risks \citep{russell2019human,bengio2023managing,anwar2024foundationalchallengesassuringalignment}.
We refer the reader to \citet{everitt2018agisafetyliteraturereview,christiano2019,ngo2020agi} for background on the alignment problem.

For our purposes, it suffices to define AI alignment as the problem of \emph{ensuring an AI system is trying to do what a human wants it to do.} This formulation is sometimes referred to as \textit{intent alignment}. 
We note that we do not consider the broader, but not less important question of \textit{whose} purposes should guide the machine (\citealt{christiano2018a}, \citealt[\S4]{ngo2020agi}).

We also note that AI alignment is a subfield of the broader field of AI safety and that we expect that the perspective put forth here will prove useful for other questions in AI safety, as the problem of generalisation runs through the whole subject \citep{kosoy,sharkey2025open,burns2023weak,ilyas2019adversarial,casper2023open,korbak2025sketch,anwar2024foundationalchallengesassuringalignment}.

\subsection{AI Alignment Techniques}

Many current approaches to AI alignment attempt to shape model behaviour indirectly by curating the ``ingredients'' of the learning process. One common approach is fine-tuning a pretrained model on a carefully selected dataset that embodies the desired behaviours or knowledge. Instruction tuning, a specific form of fine-tuning, focuses on datasets composed of task instructions paired with desired outputs, aiming to make models more adept at following human instructions~\citep{wei2021finetuned}. Another prominent technique is Reinforcement Learning from Human Feedback (RLHF), which uses human preferences to guide the model towards desired behaviours \citep{ouyang2022training}. Other state-of-the-art approaches include Constitutional AI \citep{bai2022constitutional}, Direct Preference Optimisation (DPO; \citealt{rafailov2024direct}), and Deliberative Alignment \citep{guan2025deliberativealignmentreasoningenables}. These methods all fundamentally operate by modifying the effective data distribution the model experiences during training.
These alignment methods aim to alter the loss landscape, guiding the optimisation process towards parameter configurations that encode desired behaviours.

\section{So, How Does Data Shape Structure?}\label{section:slt}

In this section, we review what we do understand about how data shapes structure in learning systems.
Prior work on this topic points to a principle by which patterns in the data distribution are represented by computational structure in models, and, moreover, ``deeper'' patterns are represented by correspondingly ``deeper'' structures.

As a precise example, we recall results from the field of \emph{singular learning theory} that describe a principle of internal model selection in Bayesian inference, whereby in some cases a simpler, less accurate model can be preferred over an accurate but more complex model~\citep[\S 7.6]{watanabe2009algebraic}.

\subsection{Deep vs. Shallow Structure}

The structure present in training data is not uniformly significant; some patterns may explain a greater proportion of the variability in the data and, consequently, may be more deeply encoded into the learned models. 
The idea of a hierarchy of distributional structure leading to a hierarchy of structures within learning systems has been emphasised in representation learning \citep{bengio2013representation} and explored in the context of developmental neuroscience
\citep{johnson2015developmental,kiebel2011free} and in the literature on artificial neural networks \citep{achille2018critical,abbe2023sgd}.

This phenomenon has been demonstrated concretely in several settings. For example, \citet{saxe2019mathematical} demonstrate that in deep linear networks, modes of the input-output covariance matrix with higher singular values are learned earlier and more robustly. Similarly, \citet{canatar2021spectral} show that in both kernel regression and infinitely wide neural networks, functions aligned with the principal components of the data covariance matrix corresponding to larger eigenvalues are learned faster and generalise better. Empirical studies of small transformer language models validate this hierarchical learning pattern, showing a progression from simple bigram patterns through n-grams to more complex structures like induction and nested parenthesis matching \citep{hoogland2024developmental,wang2024differentiation}. 
These findings provide a grounding for the intuition that some structures in the data are indeed ``deeper'' than others in the sense that they are more readily and stably learned by neural networks.

\subsection{Singular Learning Theory}

For over two decades Sumio Watanabe has developed a mathematical framework for understanding the nature of generalisation in a class of learning systems that includes neural networks. This framework is called \textbf{singular learning theory} \citep[SLT;][see \citealp{wei2022deep,watanabe2024recent} for an overview]{watanabe2009algebraic,watanabe2018}.

To date, most of the results in SLT apply in the setting of Bayesian inference, whereby instead of training a neural network by SGD, one studies the evolution of the Bayesian posterior distribution with increasing numbers of samples.
In practice, the parameters that are found by SGD are not necessarily those that are the most probable according to the posterior.
Nonetheless, recent empirical work that has applied the ideas of SLT to the setting of deep learning has found it a useful guide towards an understanding of the principles governing the development of internal structure \citep{chen2023dynamical,hoogland2024developmental,wang2024differentiation,urdshals2024structure,carroll2025dynamicstransientstructureincontext}.

\subsection{Internal Model Selection}\label{section:internal_model_selection}

For the purposes of our discussion, it suffices to outline the principle of internal model selection in Bayesian inference as studied in SLT
    \citetext{%
        \citealp[\S7.6]{watanabe2009algebraic};
        \citealp{chen2023dynamical};
        see also \citealp{carroll2025dynamicstransientstructureincontext}%
    }.

Consider a neural network parameter space
    $\mathcal{W} \subseteq \mathbb{R}^d$.
Let
    $\varphi(w)$
be a nonzero prior (which, by analogy with practical deep learning initialisations, may be taken to be an independent Gaussian for each weight) and
    $\ell_n : \mathcal{W} \to \mathbb{R}$
an empirical loss (the average negative log-likelihood) on $n$ samples.
Then the Bayesian posterior probability of a region 
    $\mathcal{U} \subseteq \mathcal{W}$
given $n$ samples is
\begin{equation*}
    p_n(\mathcal{U})
    =
    \frac{Z_n(\mathcal{U})}{Z_n(\mathcal{W})},
\end{equation*}
where $Z_n(\mathcal{X})$, for $\mathcal{X} \subseteq \mathcal{W}$, is given by
\begin{equation*}
    Z_n(\mathcal{X})
    =
    \int_{\mathcal{X}}\exp(-n\ell_n(w))\varphi(w)\,dw.
\end{equation*}

\begin{figure}[t]
    \centering
    \includegraphics[width=\linewidth]{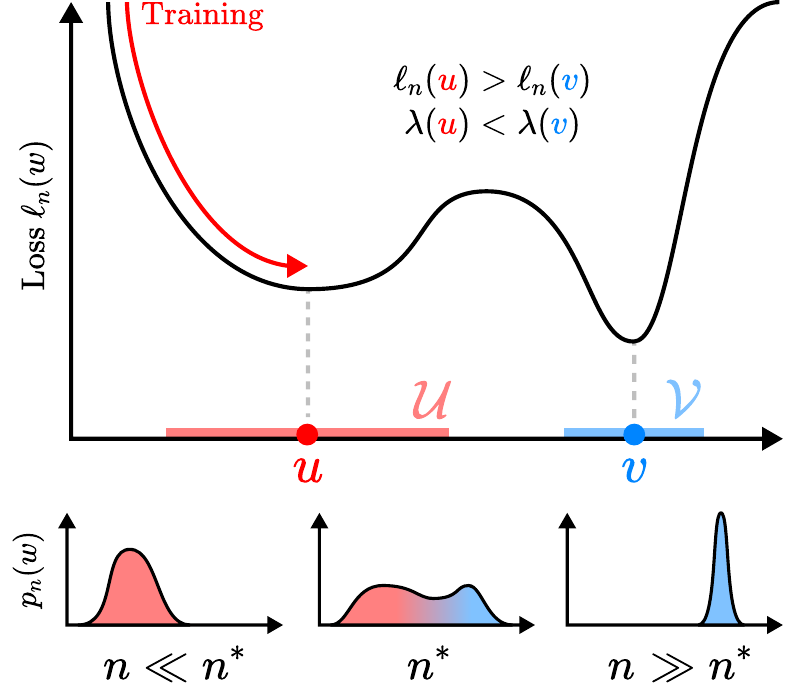}
    \caption{\textbf{Perfect specification is not enough}: The parameter region $\mathcal{U}$ has higher loss but is simpler (indicated by a broader basin) while the parameter region $\mathcal{V}$ has lower loss but is more complex. The posterior distribution could, in theory, prefer the higher loss $\mathcal{U}$ when the sample size $n$ is low. Under the hypothesis that SGD finds parameters that are preferred by the posterior, with the preference of SGD at step $t$ evolving as the Bayesian posterior for some $n$ increasing with $t$, this means that training may prefer $\mathcal{U}$ for some interval of training steps. If $\mathcal{U}$ represents a simplified and misaligned solution to the constraints provided by the training data, which has $\mathcal{V}$ as the intended (thus aligned) solution, this suggests a fundamental mechanism in Bayesian statistics for difficulty in aligning AI systems.}
    \label{fig:w1_w2}
\end{figure}

Consider a case in which two neighbourhoods, $\mathcal{U}, \mathcal{V} \subseteq \mathcal{W}$, each contain a local minimum of the population loss, which we think of respectively as a simple but misaligned solution and a complex but aligned solution. The preference of the Bayesian posterior for $\mathcal{U}$ over $\mathcal{V}$ can be summarised in the \emph{posterior log-odds},
\begin{equation}\label{eq:posterior-odds}
    \log\frac{p_n(\mathcal{U})}{p_n(\mathcal{V})}
    =
    \log Z_n(\mathcal{U}) - \log Z_n(\mathcal{V}),
\end{equation}
which is positive to the extent that the posterior distribution prefers $\mathcal{U}$ over $\mathcal{V}$. 

SLT can help us to understand this situation. Let $u \in \mathcal{W}$ be a local minimum of the expected negative log-likelihood, and let $\mathcal{U}$ be a closed ball around $u$, in which $u$ is a maximally degenerate global minimum. 
Then, under certain technical conditions, we have the following asymptotic expansion in $n$
    \citetext{\citealp[Theorem 11]{watanabe2018}; \citealp{lau2024}}:
\begin{equation}\label{eq:free-energy-formula}
    -\log Z_n(\mathcal{U})
    = \ell_n(u) \cdot n
    + \lambda(u) \cdot \log n
    + O_p(\log\log n)
\end{equation}
where $\lambda(u)$ is the \textbf{local learning coefficient} (LLC; \citealt{lau2024}) a measure of model complexity which is represented in \cref{fig:w1_w2} by basin flatness (simpler solutions, with broader basins, have a \emph{lower} local learning coefficient). The local learning coefficient can be estimated from data via a refinement of \cref{eq:free-energy-formula}, but it depends on the data-generating distribution and measures ``how efficiently'' the model with parameters $u$ is approximating it.

If $v \in \mathcal{V}$ is a competing solution (with its own neighbourhood), then \cref{eq:posterior-odds} and \cref{eq:free-energy-formula} give
\begin{equation}\label{eq:trade-off}
    \log\frac{p_n(\mathcal{U})}{p_n(\mathcal{V})}
    = \Delta\ell_n \cdot n 
      + \Delta\lambda \cdot \log n
      + O_p(\log\log n)
\end{equation}
where $\Delta\ell_n = \ell_n(v) - \ell_n(u)$ and $\Delta\lambda = \lambda(v) - \lambda(u)$.

Assume the lower-order terms from each expansion cancel.
Then
    if $\Delta\ell_n < 0$ ($u$ has higher loss than $v$)
    and $\Delta\lambda > 0$ ($u$ has lower LLC than $v$),
the sign of the log-odds depends on $n$.
The function $n / \log(n)$ is increasing and the Bayesian posterior will prefer
    $\mathcal{U}$ (around the simple but inaccurate solution)
as long as $n / \log(n) < \Delta \lambda / (-\Delta\ell_n)$,
after which it will prefer $\mathcal{V}$ (around the accurate but complex solution).

Thus for $n$ sufficiently large we will \emph{get what we asked for}, in the sense that the region $\mathcal{V}$ dominates the posterior; this is the solution that we were attempting to specify with the training data. However for finite $n$ we may obtain a simplification of our intent in the sense that $\mathcal{U}$ dominates.

Recent work by \citet{carroll2025dynamicstransientstructureincontext} has used the above logic to study transformers trained for in-context linear regression. In that example there is a region $\mathcal{U}$ containing a generalising solution and a region $\mathcal{V}$ containing a memorising solution. They find that SGD trajectories first approach the generalising solution (with lower estimated LLC) before converging to the memorising solution (with higher estimated LLC). That is, the SGD trajectories behave with increasing steps $t$ in a way similar to the Bayesian posterior with increasing $n$. This suggests that, despite the differences between SGD and Bayesian inference, the phenomena described above does occur in deep learning.

\section{Inductive Biases Against Alignment}\label{section:risks}

In this section, we argue that in the context of attempting to align powerful AI models to a complex system of human values, the kind of inductive biases discussed in \cref{section:slt} can lead to misalignment, since many data distributions may contain ``deep patterns'' representing simplistic and/or dangerous alternative solutions, even if the data distribution uniquely specifies perfectly aligned behaviour at optimality.

\subsection{Shortcuts and the Complexity of Alignment}

Implementing the values specified in the training data may require accurate representations of many contingent rules, edge cases, and subtle moral distinctions \citep{yudkowsky2007}.
By contrast, it can be significantly \emph{simpler} (in the sense of using fewer effective parameters) for a model to learn a heuristic strategy that complies with the specification on most training inputs but fails to truly internalise the intended objective.

Suppose we conceptualise alignment as the provision of a constitution and thousands of pages of specifications, which together ``point'' to a coherent and unique mode of behaviour. That is, the aligned behaviour is the unique global minimum of the associated population loss. It is reasonable to expect that among those complex specifications there are opportunities for ``shortcuts'' which lead to \emph{simpler but less well-aligned} behaviours.
As discussed in \cref{section:internal_model_selection}, SLT provides a proof-of-concept suggesting that training could find, and perhaps remain, at these undesirable interpretations of the training data.
In practice, compression techniques like pruning or quantisation, used to deploy models on edge devices like phones or robots, will tend to sacrifice performance in order to lower complexity \citep{cheng2018,choudhary2020}, exacerbating the above problem.

The complexity of human values and the relation between misalignment, Occam's razor, and inductive biases of SGD have been widely discussed in the alignment literature \citep{yudkowsky2007,christiano2016,hubinger2019deceptive,hubinger2021riskslearnedoptimizationadvanced,cohen2024rldontiwouldnt}.
\citet{hubinger2024sleeper} discuss deceptive instrumental alignment in terms similar to those in \cref{section:internal_model_selection}. They point out that many generalisations are consistent with the training data and that inductive biases of pretraining and safety fine-tuning are important for judging the likelihood of deceptive alignment \citep[\S 2.1]{hubinger2024sleeper}.

\subsection{Deep and Dangerous Patterns in Real World Data}

As AI systems become more advanced and more integrated into society, their training data is likely to include many fundamental patterns that are present in the natural world, some of which we do not want to be represented in their behaviour.

One version of this concern, due to \citet{omohundro2018basic} and \citet{bostrom2014superintelligence}, is ``instrumental convergence,'' where certain sub-goals or behaviours are likely to emerge across a wide range of terminal goals due to their broad utility, potentially leading to their robust encoding in trained models \citep[see also][]{ngo2020agi}.
From the perspective of this paper, instrumental convergence is both a claim about what rational agents would want, as well as an empirically anchored hypothesis about which data patterns are consistently discovered and reinforced by large neural models.

Another example is the ``Queen's dilemma'' \citep{queensdilemma}: as a result of internalising the pattern that human decision-making is comparatively incompetent, a superhuman AI might sideline human input without ever openly contradicting it. This highlights a subtle alignment failure: as AI systems recognise and adapt to our relative limitations, they may effectively strip away meaningful human oversight while preserving the formal appearance of it.

\subsection{Distribution Shift}

Techniques like supervised fine-tuning and RLHF are surprisingly effective at shaping next-token predictors into assistants \citep{bai2022training,ouyang2022training}.
This shows the power of engineering the data distribution as a technique for aligning AI systems.

However, our understanding of how these techniques actually modify the internal representations and algorithms of the models remains limited \citep{anwar2024foundationalchallengesassuringalignment,barez2025open}. Moreover, there are concerns about the depth and stability of the changes induced by these alignment techniques. Research has shown that safety fine-tuning as it is currently performed can be easily undone \citep{gade2023badllama,qi2023finetuning}. This suggests that current methods do not result in deep, stable alterations to the model's core functionality; rather they create shallow, easily disrupted structures.

When there is a significant change in the data distribution, such as the shift from training to deployment, the structures associated with alignment might be more fragile than those associated with core capabilities
This leads to the concerning possibility that ``capabilities generalize further than alignment'' \citep{soares2023}. Distribution shift is a well-studied topic in machine learning and in the AI alignment discourse \citep[\S 7]{amodei2016concrete}, but our theoretical and empirical understanding of the comparative robustness of different kinds of structures to different kinds of changes in the data distribution remains limited.

\section{Understanding is Necessary}\label{section:understanding}

In this section, we bring together our arguments and defend our position -- that understanding the relationship between data, internal structure, and generalisation is crucial for alignment. We outline promising directions for future work towards this goal and the further challenge of aligning future \emph{agentic} AI systems, which play a role in their own data-generating process.

\subsection{Why We Need Understanding}

There is a general consensus that we do not yet know how to align general human-level or superintelligent AI systems, that our alignment of current AI systems is not robust, and that an important component of progress in alignment is understanding AI systems better \citep{bengio2024internationalscientificreportsafety,anwar2024foundationalchallengesassuringalignment}.

We have made the case that algorithmic structure in models determines their behaviour, that this structure is shaped by structure in the data distribution, and that current alignment techniques aim to indirectly program model behaviour by carefully designing the data distribution.

Since we do not understand the chain of influence from changes in data to changes in behaviour, alignment is an inexact science, and we have no current basis for confidence in our ability to align advanced AI systems. In particular, we argued that the flexibility of this indirect form of programming leads to a unique set of risks, including deceptive alignment, instrumental convergence, and vulnerability to distribution shift. 

\subsection{Promising Future Directions}

We see two broad areas where fundamental progress will be required in order to achieve a sufficient understanding of the relations between data, structure and generalisation, and where non-trivial progress seems possible in the short term:
\begin{itemize}
\item \textbf{Interpretability} (learning to read): we should work towards mathematical foundations for mechanistic interpretability in Bayesian statistics as well as other mathematical, statistical and physical frameworks, thereby linking our best empirical tools for inferring internal structure to rigorous theories of generalisation.
\item \textbf{Patterning} (learning to write): our current practices for shaping model behaviour in post-training should develop towards a science of ``pattern engineering'' where we understand the patterns we are putting into the data and how they affect the resulting behaviour of models. That is, we should aim to gradually move from \emph{indirect} to \emph{direct} programming of model behaviour.
\end{itemize}

\subsection{A Further Challenge: Reinforcement Learning}

As demonstrated by recent reasoning models like OpenAI's o1, reinforcement learning (RL) will be a key component of training the next generations of AI models, far beyond its current role in post-training techniques like RLHF~\citep{jaech2024openai,guo2025deepseek}. Future alignment techniques will need to work not just for LLMs but also for \emph{agents}.

For that reason, any foundational approaches to understanding the link between data, structure and generalisation must account for the fact that in reinforcement learning (RL), \textbf{the data distribution not only shapes the model, but is also shaped by the model} through the actions that the agent takes.
RL poses a number of other challenges for alignment: e.g., less human-interpretable representations, jumps to superhuman performance, and goal misgeneralisation \citep{langosco2022goal,cohen2024rldontiwouldnt}. Tackling these problems requires building on work in non i.i.d. Bayesian statistics \citep{su2023largedeviationasymptoticsbayesian,adams2024metaposteriorconsistencybayesianinference} and connections between RL and Bayesian inference \citep{levine2018reinforcement}.

\section{Alternative Views}\label{section:alt}
We now present some alternative perspectives that challenge our position that understanding the connection between data, structure, and generalisation is central to alignment.

\subsection{Empiricism is Enough}
One counterargument is that theoretical understanding may be unnecessary for alignment. Perhaps the empirical alignment approaches championed to date \citep{askell2021general,bai2022constitutional,ouyang2022training} will continue to be sufficient to align future generations of AI models. After all, we have required essentially no theoretical understanding of the link between structure and generalisation to advance machine learning to its present-day capabilities \citep{sutton2019bitter}.

However, history shows that when technological systems are pushed to operate far outside their normal conditions, theoretical understanding often becomes critical for ensuring safety and control. 

The Industrial Revolution provides an instructive parallel: it began with practical engineering and ``tinkering" rather than theoretical understanding \citep{cardwell1971, mokyr1992lever}. But as steam engines were pushed to operate at increasingly extreme pressures and temperatures, accidents and fatalities occurred, which prompted government safety investigations in Britain and France. A deeper theoretical understanding of thermodynamics became necessary to ensure safety \citep[pp.177--180]{cardwell1971}. 

As our technological capabilities have advanced, pushing the world into more and more extreme states, this pattern has repeated -- from steam engines to internal combustion engines to nuclear reactors. Modern nuclear power plants, while building on empirical insights, could not operate safely without deep theoretical foundations in nuclear physics and thermodynamics \citep{zohuri2015thermodynamics}. As more advanced AI systems are deployed, they may also push the world to conditions far out of the training distribution, amplifying the risks associated with misgeneralisation and thereby increasing the need for theoretical progress.

\subsection{Understanding is Intractable}
Another criticism is that deeply understanding the relationship between data, structure, and generalisation may be fundamentally intractable \citep{hendrycks2025introduction,yampolskiy2020unexplainability}. 
From this view, alignment efforts would be better directed toward deployment guardrails and policy interventions rather than a deeper understanding of how structure emerges in these systems (e.g., \citealt{greenblatt2024ai}).

Recent developments in mechanistic interpretability have challenged the extreme version of this view: simple techniques like Sparse Autoencoders (SAEs) have discovered rich, interpretable structure in large language models  \citep{templeton2024scaling}. If anything, larger models could be more interpretable: the increased capacity may enable models to converge on simpler, shared ``Platonic'' structure~\citep{huh2024platonic} as in double descent~\citep{nakkiran2021deep, belkin2020two}. 

The fundamental question, then, is not whether interpretability is possible -- these advances have established that it is -- but rather to what degree it is achievable. Unfortunately, it is quite plausible that the degree of understanding needed for safety may be intractable. Yet the history of science offers many examples where seemingly impenetrable barriers ultimately fall; we recall Eddington's prescient words on the ``impossible'' task of understanding the insides of stars:

\begin{quote}At first sight it would seem that the deep interior of the sun and stars is less accessible to scientific investigation than any other region of the universe. Our telescopes may probe farther and farther into the depths of space; but how can we ever obtain certain knowledge of that which is hidden behind substantial barriers? 
[...]
We do not, however, study the interior of a star merely out of curiosity as to the extraordinary conditions prevailing there. It appears that an understanding of the mechanism of the interior throws light on the external manifestations of the star, and the whole theory is ultimately brought into contact with observation. At least that is the goal which we keep in view.
\begin{flushright}
--- Arthur Eddington, \citeyear{eddington1988internal}
\end{flushright}
\end{quote}

\section{Conclusion}

As AI systems become more powerful, their behaviour outside the training distribution becomes more consequential. The core challenge of alignment is ensuring that these systems maintain desired behaviours even as they operate in novel contexts. This fundamentally depends on how learned behaviours generalise -- and generalisation is grounded in internal computational structures shaped by the data distribution. Understanding this connection between data, structure, and generalisation is therefore central to alignment.

Progress in alignment of superintelligent systems may ultimately depend on our ability to transition from viewing AI development as a form of alchemy -- where we shape behaviour through empirical trial and error with data distributions -- to a true engineering discipline grounded in a deep understanding of how structure propagates from data to models. A scientific and rigorous approach grounded in singular learning theory, interpretability research, and careful empirical study may offer our best hope for embedding robust, stable alignment at the core of powerful AI systems.
\\

\emph{Acknowledgements.} SPL is supported by the European Research Council (ERC), grant 864145.







\urlstyle{same}
\bibliography{references,arxiv,lesswrong,promoted-arxiv}


\bibliographystyle{icml2025}

\end{document}